\pdfoutput=1
\documentclass[letterpaper, 10 pt, journal, twoside]{IEEEtran} 
\usepackage{graphicx}
\usepackage{lineno,hyperref}
\hypersetup{hidelinks}
\hypersetup{draft}
\usepackage[dvipsnames]{xcolor}
\modulolinenumbers[5]
\usepackage{graphics} 
\usepackage{graphicx}
\usepackage{caption}
\usepackage{subcaption}
\usepackage[ruled,vlined,linesnumbered,noresetcount]{algorithm2e}
\usepackage[]{units}
\usepackage{pgfplots}
\usepackage{tikz}
\usetikzlibrary{positioning}
\usetikzlibrary{shapes.geometric}
\usetikzlibrary{shapes.misc}
\usetikzlibrary{external}
\usepackage{paralist}
\usepackage{mathtools}

\DeclarePairedDelimiter\floor{\lfloor}{\rfloor}
\usepackage{xcolor}
\usepackage{siunitx}
\usepackage{multirow}
\usepackage{epstopdf}
\usepackage{todonotes}
\graphicspath{ {Pictures/} }
\usepackage{caption}
\usepackage{epstopdf}
\usepackage{tikz}
\usepackage{pgfplots}
\usetikzlibrary{shapes,arrows,patterns}
\usetikzlibrary{arrows.meta}
\usetikzlibrary{plotmarks}
\usetikzlibrary{calc,decorations.markings,math}
\usetikzlibrary{external}
\pgfplotsset{compat=1.14}
\usepackage{grffile}
\usepackage{bm}
  \usepgfplotslibrary{patchplots}
\usepackage{epsfig}
\usepackage{float}
\usepackage{caption}
\usepackage{amsmath} 
\usepackage{amssymb}  
\usepackage{mathtools}
\graphicspath{ {Pictures/} }
\DeclareGraphicsExtensions{.pdf,.eps}

\usepackage[acronym,nomain,nonumberlist]{glossaries}
\newacronym{admm}{ADMM}{Alternating Direction Method of Multipliers}
\newacronym{ahe}{AHE}{Adaptive Histogram Equalization}
\newacronym{ar}{AR}{Augmented Reality}
\newacronym{arw}{ ARW}{Aerial Robotic Worker}
\newacronym{ccpp}{C-CPP}{Collaborative Coverage Path Planning}
\newacronym{clahe}{CLAHE}{Contrast Limited Adaptive Histogram Equalization}
\newacronym{cnn}{CNN}{Convolutional Neural Network}
\newacronym{cpp}{CPP}{Coverage Path Planning}
\newacronym{dbscan}{DBSCAN}{Density-Based Spatial Clustering of Applications with Noise }
\newacronym{dcm}{ DCM }{ Direct Cosines Matrix }
\newacronym{dl}{DL}{Deep Learning}
\newacronym{dof}{DoF}{Degree of Freedom}
\newacronym{ekf}{EKF}{Extended Kalman Filter}
\newacronym{ensc}{ENSC}{ Elastic Net Subspace Clustering}
\newacronym{fbe}{FBE}{Forward Backward Envelope}
\newacronym{fov}{FoV}{Field of View}
\newacronym{fps}{fps}{Frame Per Second}
\newacronym{ga}{GA}{Genetic Algorithm}
\newacronym{gdp}{GDP}{Gross Domestic Product}
\newacronym{gnss}{GNSS}{Global Navigation Satellite System}
\newacronym{gps}{GPS}{Global Positioning System}
\newacronym{iid}{IID}{Independent Identically Distributed}
\newacronym{imu}{IMU}{Inertial Measurement Unit}
\newacronym{kf}{KF}{Extended Kalman Filter}
\newacronym{llwcf}{LLWCF}{Left Local Wall Continuum Function}
\newacronym{lof}{LOF}{Local Outlier Factor}
\newacronym{lrcn}{LRCN}{ Long-Term Recurrent Convolutional Network}
\newacronym{lwcf}{LWCF}{Local Wall Continuum Function}
\newacronym{mae}{MAE}{Mean Absolute Error}
\newacronym{mav}{MAV}{Micro Aerial Vehicle}
\newacronym{mhe}{MHE}{Moving Horizon Estimation}
\newacronym{mimo}{MIMO}{Multiple Input Multiple Output}
\newacronym{mlp}{MLP}{MultiLayer Perceptron}
\newacronym{mpc}{MPC}{Model Predictive Control}
\newacronym{msf}{MSF}{Multi Sensor Fusion}
\newacronym{nmhe}{NMHE}{Nonlinear Moving Horizon Estimation}
\newacronym{nmpc}{NMPC}{Nonlinear Model Predictive Control}
\newacronym{nn}{NN}{Nueral Network}
\newacronym{nwu}{NWU}{North-West-Up}
\newacronym{open}{OpEn}{Optimization Engine}
\newacronym{panoc}{PANOC}{Proximal Averaged Newton-type method for Optimal Control}
\newacronym{pca}{PCA}{Principal Component Analysis}
\newacronym{pdf}{PDF}{Probability Density Function}
\newacronym{pid}{PID}{Proportional Integral Derivative}
\newacronym{ps}{PS}{Pattern Search}
\newacronym{psd}{PSD}{ Positive Semi-Definite }
\newacronym{pso}{PSO}{Particle Swarm Optimization}
\newacronym{ransac}{RANSAC}{Random sample consensus}
\newacronym{rbf}{RBF}{ Radial Basis Function}
\newacronym{relu}{ReLU}{Rectified Linear Unit}
\newacronym{rl}{RL}{Reinforcement Learning}
\newacronym{rlwcf}{RLWCF}{Right Local Wall Continuum Function}
\newacronym{rmse}{RMSE}{Root Mean Square Error}
\newacronym{ros}{ROS}{Robot Operating System}
\newacronym{sfm}{SfM}{Structure from Motion}
\newacronym{slam}{SLAM}{Simultaneous Localization and Mapping}
\newacronym{slap}{SLAP}{Simultaneous Localization and Planning}
\newacronym{spam}{SPAMS}{SPArse Modeling Software}
\newacronym{sqp}{SQP}{Sequential Quadratic Programming}
\newacronym{ssc-admm}{SSC-ADMM}{ Sparse Subspace Clustering Alternating Direction Method of Multipliers}
\newacronym{ssc-omp}{SSC-OMP}{Sparse Subspace Clustering Orthogonal Matching Pursuit}
\newacronym{uav}{UAV}{Unmanned Aerial Vehicle}
\newacronym{uaw}{ UAW}{Unmanned Aerial Workers}
\newacronym{ugv}{UGV}{Unmanned Ground Vehicle}
\newacronym{ukf}{UKF}{Unscented Kalman Filter}
\newacronym{uos}{UoS}{Union of Subspaces}
\newacronym{upac}{UPAC}{Utilities, Patrol and Construction Committee}
\newacronym{vi}{VI}{Visual Inertia}
\newacronym{vo}{VO}{Visual Odometry}
\newacronym{zepa}{ZEPA}{Zero Entry Production Areas}

\usepackage{url}
\newlength\fwidth

\author{Sina Sharif Mansouri$^{1}$, Christoforos Kanellakis$^{1}$, Bj\"orn Lindqvist$^{1}$, Farhad Pourkamali-Anaraki$^{2}$, \\ Ali-akbar Agha-mohammadi$^{3}$, Joel Burdick$^{4}$ and George Nikolakopoulos$^{1}$ 
\thanks{Manuscript received February 24, 2020; revised June 1, 2020; accepted June 29, 2020.}
\thanks{This paper was recommended for publication by Editor N. Amato upon evaluation of the Associate Editor and Reviewers' comments.
This work has been partially funded by the European Unions Horizon 2020 Research and Innovation Programme under the Grant Agreement No. 730302 SIMS.} 
\thanks{$^{1}$The authors are with the Robotics Team, Department of Computer, Electrical and Space Engineering, Lule\r{a} University of Technology, Lule\r{a} SE-97187, Sweden. First Author's email: \texttt{sinsha@ltu.se}.}
\thanks{$^{2}$The author is with  Department of Computer Science, University of Massachusetts Lowell, MA, USA. Fourth Author's email: \texttt{farhad\_pourkamali@uml.edu}.}%
\thanks{$^{3}$The author is with Jet Propulsion Laboratory California Institute of
Technology Pasadena, CA, 91109. Fifth Author's email: \texttt{aliagha4@gmail.com}.}%
\thanks{$^{4}$The author is with Division of Engineering and Applied Sciences, California Institute of Technology, Pasadena, California, USA. Sixth Author's email: \texttt{jwb@robotics.caltech.edu}.}%
\thanks{Digital Object Identifier (DOI): see top of this page.}
}

\title{A Unified NMPC Scheme for MAVs Navigation with 3D Collision Avoidance under Position Uncertainty}

\begin{document}
\maketitle              

\begin{abstract}
This article proposes a novel \gls{nmpc} framework for \gls{mav} autonomous navigation in indoor enclosed environments. 
The introduced framework allows us to consider the nonlinear dynamics of \gls{mav}s, nonlinear geometric constraints, while guarantees real-time performance. 
Our first contribution is to reveal underlying planes within a 3D point cloud, obtained from a 3D lidar scanner, by designing an efficient subspace clustering method.
The second contribution is to incorporate the extracted information into the nonlinear constraints of \gls{nmpc} for avoiding collisions. 
Our third contribution focuses on making the controller robust by considering the uncertainty of localization in \gls{nmpc} using Shannon's entropy to define the weights involved in the optimization process. 
This strategy enables us to track position or velocity references or none in the event of losing track of position or velocity estimations. 
As a result, the collision avoidance constraints are defined in the local coordinates of the \gls{mav} and it remains active and guarantees collision avoidance, despite localization uncertainties, e.g., position estimation drifts. 
The efficacy of the suggested framework has been evaluated using various simulations in the Gazebo environment. 


\end{abstract}

\glsresetall
%
%
%
\section{Introduction}
\IEEEPARstart{T}{he} deployment of \glspl{mav} is gaining more attention in different applications for interior or exterior inspections. Examples include infrastructure inspection~\cite{mansouri2018cooperative}, underground mine tunnel inspection~\cite{mansouri2020deploying, mansouri2020nmpcmine}, and bridge inspection~\cite{hallermann2014visual}. The obtained information from the \gls{mav} inspections can be used for identifying various types of damages, while providing real-time map-building and navigation in unknown and complex environments. 

One of the main challenges in deploying \gls{mav}s in real-world applications is the ability to provide collision-free paths, which strongly depend on an accurate and robust localization. A failure in the localization adversely impacts the overall mission quality, and in the worst-case scenario, it leads to a collision/crash. Unfortunately, localization uncertainties are inevitable in \gls{gps}-denied environments and in areas that lack prominent visual and geometric features. 
Hence, there is a need to develop new methods to handle such uncertain cases in perceptually-degraded and extreme environments such as subterranean environments, to provide access to unreachable areas and increase the personnel's overall safety. 

This article proposes a new \gls{nmpc} architecture for path planning and collision avoidance that considers localization uncertainties as well as geometrically-induced constraints. 
In the proposed method, a novel and efficient subspace clustering technique~\cite{parsons2004subspace} will be presented for the transformation of geometric constraints into equivalent plane constraints for modeling the surrounding environment. The proposed approach significantly reduces the computational complexity and memory usage of existing subspace clustering methods. 
Due to the real-time constraints, traditional subspace clustering methods are not applicable, as the computational time is often greater than one second, even for one thousand data points (e.g.,~\cite{you2016scalable}). 
The extracted equations are in the body frame and are thus decoupled from localization and used in the sequel as nonlinear constraints in \gls{nmpc} for avoiding collisions. 
Additionally, to cope with uncertainties in the localization, we assume that the weights of the trajectory following in the \gls{nmpc} are adaptive and vary based on Shannon's entropy~\cite{shannon1948mathematical} of the measurements. 
This feature permits the \gls{mav} to carry out the mission despite significant uncertainties in localization by dropping one or both of the position or velocity references from tracking objectives.
Hence, even a failure in the high level path planner or localization will not lead to a collision of the platform. 
The proposed control architecture allows for continued mission by \gls{mav} even in the presence of localization drift. 



\subsection{Background \& Motivation}
In the related literature, there have been many works that addressed control, navigation and path planning of \gls{mav}s~\cite{lavalle2006planning, mo2019nonlinear}. Most of these works decouple the problems of designing controllers, navigation and path planning, and proposed hierarchical structures for autonomous navigation. 

Towards the topic of \gls{mav} navigation and path planning, exploration algorithms like frontier exploration algorithms~\cite{fentanes2011new}, entropy based algorithms~\cite{burgard2005coordinated}, and information-gain algorithms~\cite{bhattacharya2013distributed} provide a global planning strategy for \gls{mav}s. These methods mainly rely on information concerning the map and localization of the platform and compute regions that reduce the map uncertainty. The obtained desired areas to visit are then fed to the controller to generate commands for navigation of \gls{mav}s~\cite{ carrillo2015autonomous, bircher2016receding}. As localization and mapping suffer from uncertainties and drifts, an additional reactive control layer is often considered for local obstacle avoidance and to prevent collisions with the environment. The most widely used reactive control layer is the artificial potential field~\cite{barraquand1992numerical}, which has seen wide use in multiple application areas, as a local path planner for both \gls{mav}s and other robotic platform applications, such as the mapping of an infrastructure~\cite{droeschel2016multilayered} or in the case of the multi-robot coordination~\cite{song2002potential} for mobile robots. However, this method suffers from getting stuck in a local optimum and the result would be conservative and not optimal~\cite{hoy2015algorithms}. Few articles have addressed collision avoidance of \gls{mav}s with the \gls{nmpc} framework. Much research focused on addressing formation problems with a fixed number of agents~\cite{ille2017collision}, considered a linear model~\cite{baca2018model}, assumed global pose information of obstacles~\cite{small_panoc_2018}, uncertainties of position estimations are excluded.


Plane segmentation has been extensively investigated, and we present a brief overview. These methods can be divided into three categories. (a) Point clustering, based on similarities between the measurements such as distance and angle between surface normal~\cite{sun2019oriented}. 
(b) Region growing, where the method chooses seed points or regions, and cluster the points based on that information. 
The authors in~\cite{zermas2017fast} proposed fast segmentation of 3D point clouds for autonomous vehicles. This method extracts a set of seed points based on low height values to estimate the ground surface and then extract the points close to initial ground plane. (c) \gls{ransac}-based plane fitting where points from the point cloud are sampled, and planar models are fitted to them. In~\cite{qian2014ncc}, a \gls{ransac} method was proposed for performing normal coherence check on points and removed the data points whose normal directions were contradictory to the fitted plane. While plane detection has received significant attention over the last years, most research relies on organized point clouds, such as RGB-D images~\cite{feng2014fast}, where the neighbor information can be used. However, extracting planes from unorganized point clouds is more challenging because of the cloud size variations, which means that the neighbor information cannot be immediately used~\cite{li2017improved}.

In machine learning, sparse subspace clustering~\cite{elhamifar2013sparse} is the state-of-the art method for segmenting points drawn from a union of subspaces. These methods consist of two steps: 1) solving the sparse representation problem to find a similarity matrix, 2) employing spectral clustering to partition the data. Previous research efforts focused solely on reducing the cost associated with step 1. Unfortunately, existing methods did not attempt to reduce the cost of spectral clustering. Therefore, state-of-the-art subspace clustering methods rely on the increase in computational power to perform step 2, which is a bottleneck for real-time data processing.

\subsection{Contributions}

The first contribution is to design a scalable subspace clustering technique that finds clusters in the form of subspaces within a 3D point cloud. We develop a scalable method for partitioning the input data into clusters using three tools: (1) randomized sampling, (2) fast sparse representation solvers, and (3) efficient methods for computing the eigenvalue decomposition. Thus, the proposed method enables extracting the required information for avoiding collisions.

The second contribution stems from coupling the 3D collision avoidance and controller in the body frame. 
The obtained plane equations in local coordinates are considered as non-linear constraints in the optimization scheme, while the controller tracks the desired trajectory in global coordinates. The proposed \gls{nmpc} is solved by \gls{open}~\cite{open2019} that uses the \gls{panoc} algorithm~\cite{sathya2018embedded}, while a penalty method is used for enforcing equality constraints that guarantees obstacle avoidance.  

The third contribution enables the framework to handle localization uncertainties by defining adaptive weights for tracking the position and velocity reference way-points. The weights are calculated based on uncertainties associated with measurements, while drifts in localization estimations result in position or velocity tracking or neither. The constraints are defined based on the local point cloud and will be active in all cases, and collision avoidance is guaranteed. The proposed solution results in progressing the \gls{mav} navigation, instead of terminating the mission or collision of the platform. 

The final contribution is to thoroughly examine the performance of the proposed method in corridor and confined environments using Gazebo. The proposed method successfully avoids collisions, even when the localization estimations are uncertain. These results show the capability of the proposed architecture in challenging scenarios and can be found in the following link:
\url{https://youtu.be/76ob9HSrOAs}

\subsection{Outline}
Section~\ref{sec:prelimanaries} introduces notations used in the article. Then, the segmentation approach is presented in Section~\ref{sec:multiplanesegmentation}. In the sequel, a presentation of the \gls{nmpc} formulation, and the solver are described in Section~\ref{sec:nmpc}. Section~\ref{sec:result} presents our simulation results in the Gazebo environment. Finally, Sections~\ref{sec:limitations} and~\ref{sec:conclusions} conclude the article by summarizing the findings and offering some directions for future research. 

\section{Notation and Preliminaries} \label{sec:prelimanaries}

\begin{figure*}[htbp!] \centering 
\includegraphics[width=0.76\linewidth]{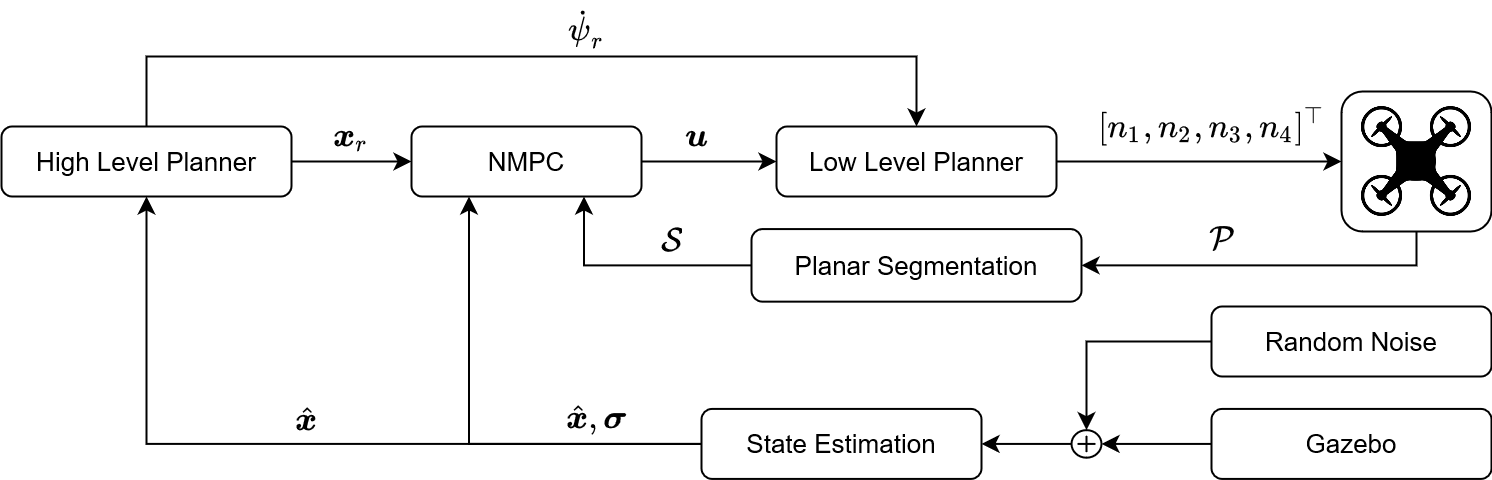}
\caption{Control scheme of the proposed \gls{nmpc} module, where way point and yaw rate references are provided from the high level controller, \gls{nmpc} generates thrust and attitude commands and the low level controller generates motor commands.}
\label{fig:Controllerscheme}
\end{figure*}

The empty set in $\mathbb{R}^n$ is denoted by $\emptyset_n$. A vector in $\mathbb{R}^n$ is predetermined as column vector in $\mathbb{R}^{n\times 1}$.
The scalar product between two vectors $\bm{r},\bm{t}\in\mathbb{R}^n$ is denoted by $\bm{r} \cdot \bm{t}$. 
The identity matrix in $\mathbb{R}^{n\times n}$ is denoted by $\mathbf{I}_n$.
The transpose of a matrix $\mathbf{M}\in \mathbb{R}^{n\times m}$ is denoted by $\mathbf{M}^\top$. 
The set $\mathcal{P}=\{(x_i,y_i,z_i), i\in \mathbb{N}  \}$ presents the set of point clouds in $\mathbb{R}^3$.
The set $\mathcal{S}=\{(\alpha_i,\beta_i,\gamma_i,\zeta_i), i\in \mathbb{N}  \}$ denotes the set of plane equation in the form of $\alpha x+ \beta y+ \gamma z+ \zeta =0, \,\alpha, \beta,\gamma,\zeta \in \mathbb{R}$.
$\bm{x}$, $\bm{u}$ are called \emph{state} and \emph{input} vectors respectively. $\bm{p}=[p_x,p_y,p_z]^\top \in \mathbb{R}^3$ is the position and $\bm{v} = [v_x, v_y, v_z]^\top \in \mathbb{R}^3$ is the vector of linear velocities, $\phi \in \mathbb{R} \cap [-\pi,\pi]$ and $\theta \in \mathbb{R}\cap [-\pi,\pi]$ are the roll and pitch angles. Figure~\ref{fig:Controllerscheme} depicts the block diagram of the proposed structure with the high level \gls{nmpc} controller, and the low level controller with the \gls{mav} in the loop. The set $\mathcal{S}$ is provided from the plane segmentation module (Section \ref{sec:multiplanesegmentation}) and the \gls{nmpc} module (Section~\ref{sec:nmpc}) generates control actions $\bm{u}$ for navigating to the reference way point $\bm{x}_r$, based on the odometry uncertainty $\boldsymbol{\sigma}$, the estimated states $\hat{\bm{x}}$, and the plane equations. 

\section{Segmentation of 3D Point Cloud} \label{sec:multiplanesegmentation}
In this section, we propose an efficient subspace clustering technique to extract planes or surfaces that are presented in a point cloud. Suppose that the obtained 3D point cloud consists of $n$ points, i.e., $\mathcal{P}=\{(x_i,y_i,z_i)\}_{i=1}^n$, then we will form a data matrix $\mathbf{B}=[\mathbf{b}_1,\ldots,\mathbf{b}_n]\in\mathbb{R}^{3\times n}$ by assuming each data point as a column vector. As these points are drawn from a union of subspaces in $\mathbb{R}^3$, we can express each point as a sparse linear combination of all the other data points. 

Subspace clustering techniques aim to partition the data into multiple clusters and fit each cluster with a low-dimensional subspace (e.g., a 2D surface). A popular approach to subspace clustering relies on  solving the following optimization problem for each point $\mathbf{b}_j$, $j=1,\ldots,n$:
\begin{align}
    \min_{\mathbf{c}_j\in\mathbb{R}^n} \|\mathbf{c}_j\|_1+\frac{\lambda}{2} \|\mathbf{b}_j - \sum_{i\neq j} c_{ij} \mathbf{b}_i\|_2^2\;\;\text{s.t.}\;\;\mathbf{c}_j^T\mathbf{1}_{n}=1,
\end{align}
where $\|\cdot\|_q$ represents the $\ell_q$ norm for vectors, $\lambda$ is a regularization parameter, $\mathbf{c}_j=[c_{1j},\ldots,c_{nj}]^T$ is the coefficient vector associated with $\mathbf{b}_j$, and $\mathbf{1}_n$ is the vector of all ones. The constraint $i\neq j$ avoids the trivial solution of expressing $\mathbf{b}_j$ via itself, and the constraint $\mathbf{c}_j^T\mathbf{1}_{n}=1$ allows us to extract the general case of affine subspaces. 

The above optimization problem can be cast in a more concise form for the entire data set, i.e., $\mathbf{b}_j,\;j=1,\ldots,n$:
\begin{subequations} 
\begin{align}
    & \min_{\mathbf{C}\in\mathbb{R}^{n\times n}} \|\mathbf{C}\|_1+\frac{\lambda}{2} \|\mathbf{B}-\mathbf{B}\mathbf{C}\|_F^2 \\
    & \text{s.t.}\;\;\text{diag}(\mathbf{C})=\mathbf{0}_n,\;\;\mathbf{C}^T\mathbf{1}_n=\mathbf{1}_n, 
    \end{align}
    \label{eq: sole}
\end{subequations}
where $\|\mathbf{C}\|_1=\sum_{i,j}|c_{ij}|$, $\|\mathbf{C}\|_F^2=\sum_{i,j}c_{ij}^2$, and $\text{diag}(\mathbf{C})$ is the vector of the diagonal elements of $\mathbf{C}$. After solving the above optimization problem and finding the coefficient matrix $\mathbf{C}=[\mathbf{c}_1,\ldots, \mathbf{c}_n]$, the next step is to find the segmentation of the data into multiple subspaces. To this end, we build a weighted graph with $n$ nodes corresponding to the $n$ original data points, and the similarity matrix is defined as $\mathbf{W}=|\mathbf{C}|+|\mathbf{C}|^T$. Finally, we apply spectral clustering to $\mathbf{W}$ for partitioning the data into subspaces. To be formal, we form the normalized graph Laplacian matrix $\mathbf{L}=\mathbf{I}_n - \mathbf{D}^{-1/2}\mathbf{W}\mathbf{D}^{-1/2}$, where $\mathbf{D}$ is the diagonal degree matrix. Then, the $K$ eigenvectors $\mathbf{v}_1,\ldots,\mathbf{v}_K\in\mathbb{R}^n$ corresponding to the $K$ smallest eigenvalues of $\mathbf{L}$ are found ($K$ represents the number of clusters). The last step of spectral clustering is to perform \textit{K-means} clustering on the rows of the matrix $\mathbf{V}=[\mathbf{v}_1,\ldots,\mathbf{v}_K]\in\mathbb{R}^{n\times K}$ to find the segmentation of the original data points. 
The optimization problem in \eqref{eq: sole} can be solved using the
\gls{admm} \cite{boyd2011distributed}, which scales poorly with the data size. The computational cost of existing implementations is cubic or quadratic in terms of the number of data points. Moreover, one has to form various $n$-dimensional square matrices (e.g., $\mathbf{C}$ and $\mathbf{W}$) that will lead to high memory usage. Hence, existing subspace clustering methods are not appropriate for our problem of interest. We address these problems by developing a scalable subspace clustering technique.

The high computational cost of constructing the coefficient matrix originates from computing a regularized representation of every single data point with respect to the whole dataset. Thus, the first step of our novel method is to form two subsets of the original data by uniform sampling without replacement. Furthermore, we eliminate the constraint $\mathbf{C}^T\mathbf{1}_n=\mathbf{1}_n$ by mapping the original data from $\mathbb{R}^3$ to $\mathbb{R}^4$. This trick is known as homogeneous embedding \cite{li2018geometric}, where we add a new coordinate which is $1$ (or another constant) to every column of the matrix $\mathbf{B}$, i.e., $\mathcal{P}=\{(x_i,y_i,z_i,1)\}_{i=1}^n$.

Given two sampling parameters $0<\kappa_1< \kappa_2<0.5$, we create two sets of indices $\mathcal{I}_1$ and $\mathcal{I}_2$ with $n_1=\floor{ \kappa_1 n}$ and $n_2=\floor{\kappa_2 n}$ elements from $\{1,\ldots,n\}$ selected uniformly at random, respectively. Then, we modify the sparse representation problem for each $\mathbf{b}_j$, $j\in\mathcal{I}_2$: $ \min_{\mathbf{c}_j\in\mathbb{R}^{n_1}} \|\mathbf{c}_j\|_1+\frac{\lambda}{2} \|\mathbf{b}_j - \sum_{i\in\mathcal{I}_1} c_{ij} \mathbf{b}_i\|_2^2$.  
This problem does not return a trivial solution because $\mathcal{I}_1\cap \mathcal{I}_2=\emptyset$, and we propose to use the \glspl{spam} package \cite{mairal2014sparse}.

After solving the new optimization problem, we should apply spectral clustering to the obtained coefficient matrix $\mathbf{C}=[\mathbf{c}_1,\ldots,\mathbf{c}_{n_2}]$. However, the matrix $\mathbf{C}$ is not square anymore because $n_1\neq n_2$, and we often want $n_1$ to be much smaller than $n_2$ to reduce the computational cost. 

State-of-the-art techniques seek to build a square matrix of size $n_2\times n_2$, which can be quite expensive. To tackle this problem, we \textit{implicitly} form the similarity matrix $\mathbf{W}=\widetilde{\mathbf{C}}^T\widetilde{\mathbf{C}}\in\mathbb{R}^{n_2\times n_2}$, where $\widetilde{\mathbf{C}}=|\mathbf{C}|$. Next, we present an efficient approach to perform spectral clustering using the new similarity matrix. We compute $i$-th element of $\mathbf{D}$ as:
\begin{align}
    \sum_{j=1}^{n_2} w_{ij}=\sum_{j=1}^{n_2} \widetilde{\mathbf{c}}_i^T\widetilde{\mathbf{c}}_j=\widetilde{\mathbf{c}}_i^T\boldsymbol{\eta}=\widetilde{\mathbf{c}}_i \cdot \boldsymbol{\eta},
\end{align}
where $\boldsymbol{\eta}=\sum_{j=1}^{n_2}\widetilde{\mathbf{c}}_j\in\mathbb{R}^{n_1}$. Thus, we compute the diagonal degree matrix $\mathbf{D}$ using $n_2$ scalar products.

The remaining task is to compute the $K$ smallest eigenvectors of the graph Laplacian matrix. We can reduce the computational cost and memory usage of this step by computing the top $K$ eigenvectors of $\mathbf{I}_n - \mathbf{L}=\mathbf{D}^{-1/2}\mathbf{W}\mathbf{D}^{-1/2}$. Let $\mathbf{U}\boldsymbol{\Sigma}\mathbf{P}^T$
be the singular value decomposition of $\widetilde{\mathbf{C}}\mathbf{D}^{-1/2}\in\mathbb{R}^{n_1\times n_2}$, where $\mathbf{U}\in\mathbb{R}^{n_1\times r}$ (left singular vectors), $\mathbf{P}\in\mathbb{R}^{n_2\times r}$ (right singular vectors), $\boldsymbol{\Sigma}$ contains the singular values, and $r$ is the rank parameter. Then, the top $K$ eigenvectors of $\mathbf{D}^{-1/2}\mathbf{W}\mathbf{D}^{-1/2}$ are equivalent to the top $K$ right singular vectors of $\widetilde{\mathbf{C}}\mathbf{D}^{-1/2}$ since we have \cite{pourkamali2019large}:
\begin{align}
    \mathbf{D}^{-1/2}\mathbf{W} \mathbf{D}^{-1/2}=(\mathbf{U}\boldsymbol{\Sigma}\mathbf{P}^T)^T(\mathbf{U}\boldsymbol{\Sigma}\mathbf{P}^T)=\mathbf{P}\boldsymbol{\Sigma}^2\mathbf{P}^T.
\end{align}
After computing the top eigenvectors and performing \textit{K-means} clustering, we obtain the segmentation of the data. The final step picks three points within each subspace to find the equation of each subspace for collision avoidance.

\section{Nonlinear Model Predictive Control} \label{sec:nmpc}
\subsection{Objective Function}

We develop the \gls{nmpc} with 3D collision avoidance constraints, and we solve it by \gls{panoc}~\cite{sathya2018embedded} to guarantee real-time performance. The objective of \gls{nmpc} is to track the reference trajectory $\bm{x}=[p,v,\phi,\theta]^\top$ from high level planner or operator and to generate thrust $T$ and attitude commands $\phi_d, \theta_d$ for a low level controller, while guaranteeing safety distance from all extracted planes. Based on $\bm{u}=[T, \phi_d, \theta_d]^\top$, and reference yaw rate command $\dot{\psi}_r$ from a high level planner, the low level controller generates motor commands $[n_1,\dots,n_4]^\top$ for the \gls{mav}. 

The states of the non-linear dynamics of the \gls{mav} based on~\cite{mav_linear_mpc} can be presented as,
\(
	  \bm{x}
    {}={}
	  [p_x,p_y,p_z,v_x,v_y,v_z,\phi,\theta]^\top
\), 
\(
	  \hat{\bm{x}}
    {}={}
	  [\hat{p}_x,\hat{p}_y,\hat{p}_z,\hat{v}_x,\hat{v}_y,\hat{v}_z,\hat{\phi},\hat{\theta}]^\top
\) 
is the estimated state from \gls{ekf} for \gls{mav} dynamics, and the control input is: 
\(
	  \bm{u}
    {}={}
	  [T, \phi_d,\theta_d]^\top
\).
The discrete-time dynamical system is obtained by Euler method and with a sampling time of $T_s$ as $
 \bm{x}_{t+1} = f(\bm{x}_t, \bm{u}_t).$

In the \gls{nmpc} approach, a finite-horizon problem with prediction horizon $N$ is solved at every time instant $k$. The states and control actions $k+j$ steps ahead of the current time step $k$ are expressed by $\bm{x}_{k+j|k}$, and $\bm{u}_{k+j\mid k}$ respectively. At each time step, \gls{nmpc} generates an optimal sequence of control actions $\bm{u}_{k|k}^{\star}$, $\dots$, $\bm{u}_{k+N-1|k}^{\star}$, and the first control action $\bm{u}_{k|k}^\star$ is applied to the flight controller using a zero-order hold element, that is, $\bm{u}(t)=\bm{u}_{k|k}^\star$ for $t\in [kT_s, (k+1)T_s]$. For the proposed \gls{nmpc}, the following finite horizon cost function is introduced:
\begin{multline} \label{eq:costfunction} 
J = \sum_{j=0}^{N-1} 
  \underbrace{\|\bm{x}_{k+j+1|k}-\bm{x}_{r}\|_{\bm{Q}_x}^2}_\text{way point error} \\     
   + \underbrace{\|\bm{u}_{k+j+1|k}-\bm{u}_{r}\|_{\bm{Q}_u}^2}_\text{actuation} 
   + \underbrace{\|\bm{u}_{k+j|k}-\bm{u}_{k+j-1|k}\|_{\bm{Q}_{\Delta u}}^2 }_\text{smoothness cost}.
\end{multline} 

The first term of $J$ gives the tracking of the reference way points by penalizing a deviation from $\bm{x}_{r}$. 
The second term is the hovering term, where $\bm{u}_{ref}$ is $[g,\, 0,\, 0]^\top$, which is the 
hover thrust with horizontal angles. 
The third term penalizes the aggressiveness of the obtained control actions. Additionally, $\bm{Q}_x \in \mathbb{R}^{8\times 8}$, $\bm{Q}_u\in \mathbb{R}^{3\times 3}$, 
$\bm{Q}_{\Delta u}\in \mathbb{R}^{3\times 3}$ are the weights for each term of the objective function, 
which reflects the relative importance of each term. To consider uncertainties in the localization estimation, the term $\bm{Q}_x$ is a diagonal matrix and elements are updated based on the Shannon entropy of the measurements. We consider set of measurements variances as:
\begin{multline}
   \bm{\sigma}=\{(\sigma^{p_x}_i,\sigma^{p_y}_i,\sigma^{p_z}_i,\sigma^{v_x}_i,\sigma^{v_y}_i,\sigma^{v_z}_i), \\
   i \in \{k-(n_{max}-1),\dots,k\}\} ,
\end{multline}
where $\sigma$ is the variance of measurements and $n_{max}$ is limited window size of stored previous variances. We compute the Shannon entropy for each set of variance at time instant $k$ by first defining  probabilities $P_i=\sigma_i/(\sum_{i=1}^{n_{max}} \sigma_i)$.
Next, we compute the Shannon entropy, i.e., $H = -\sum_{i=1}^{n_{max}} P_i \log(P_i)$.
Then, the obtained $H$ will be used as diagonal terms of $\bm{Q}_x$ as $\bm{Q}_x=diag(H^{p_x},H^{p_y},H^{p_z},H^{v_x},H^{v_y},H^{v_z})$


\subsection{Obstacle Definition and Constraints}

\subsubsection{Collision Avoidance Constraint}
The set of extracted planes are presented by:
\begin{equation}
    \mathcal{S}=\{(\alpha_i,\beta_i,\gamma_i,\zeta_i), i=1,\dots,n_{cluster}  \},
\end{equation}
where $n_{cluster}$ is the number of clusters and $\alpha$, $\beta$, $\gamma$, $\zeta$ are the plane coefficients. To avoid the obstacles, for each extracted plane equation, the constraints are defined in~\eqref{eq:constraintscollision}. It should be highlighted that the constraints are defined in the body frame ($\bm{p}_{k+j|k}-\bm{p}_{k|k}$) and the \gls{mav} is considered in the center of the point cloud so that the global position is not required.
\begin{equation} \label{eq:constraintscollision}
 d_{s} \le  \frac{|[\alpha_i, \beta_i, \gamma_i]\cdot[\bm{p}_{k+j|k}-\bm{p}_{k|k}]^\top + \zeta_i|}{\sqrt{\alpha_i^2+\beta_i^2+\gamma_i^2}} ,  
\end{equation}
for $i = 0, \ldots, n_{cluster}$ and $j = 0, \ldots, N-1$. The proposed constraints guarantee that the \gls{mav} has at least $d_s$ distance to each plane. 

\subsubsection{Input Constraint}
To prevent the aggressive behavior of control actions, the following input constraints on the successive differences of control actions are defined as:
\begin{subequations} \label{eq:inputdelta_constraints}
\begin{align}
    |\phi_{d,k+j|k} - \phi_{d,k+j+1|k}| 
    {}\leq{}
    \Delta \phi_{\max},
    \\
    |\theta_{d,k+j|k}  - \theta_{d,k+j+1|k}| 
    {}\leq{}
    \Delta \theta_{\max},
\end{align}
\end{subequations}
for $j = 0, \ldots, N-1$. Where $\Delta \phi_{\max}$ and $\Delta \theta_{\max}$ are bounds for changes. 
\subsection{Embedded Optimization}
The following optimization problem can be defined:
\begin{subequations} \label{eq:nmpc}
\begin{align}
&\underset{
  \{  \bm{u}_{k+j\mid k} \}_{j=0}^{N-1}} {\mathbf{minimize}}\  J \\
&\mathbf{subject\,to}:\bm{x}_{k+j+1\mid k}=f(\bm{x}_{k+j\mid k},\bm{u}_{k+j\mid k}), 
\label{eq:optimizationcoverage:sysDyn}
\\
& \text{Constraints~\eqref{eq:constraintscollision},~\eqref{eq:inputdelta_constraints}},\label{eq:optimizationcoverage:constraint}
\\ 
&\bm{u}_{k+j\mid k} \in [\bm{u}_{\min},\bm{u}_{\max}],
\label{eq:optimizationcoverage:actuationConstraints}
\end{align}
\end{subequations}
for $j=0,\ldots, N-1$.

Problem \eqref{eq:nmpc} is a parametric non-convex optimization problem that must be 
solved online. 
We use the fast optimization solver Optimization Engine (for short \texttt{OpEn}) developed by~\cite{open2019}. 
We first eliminate the state sequence following the procedure detailed in~\cite{sathya2018embedded}. Hence, we have a problem of the decision variable $\bm{u}=(\bm{u}_{k\mid k}, \bm{u}_{k+1\mid k},\ldots,\bm{u}_{k+N-1\mid k})\in\mathbb{R}^{3N}$,
which is constrained in $\bm{U}=[\bm{u}_{\min}, \bm{u}_{\max}]^{N}$. For the $i^{th}$ plane equation,
constraint \eqref{eq:constraintscollision} reduces to:
\begin{equation}
 \max\left\{0, d_{s} -  \frac{|[\alpha_i, \beta_i, \gamma_i]\cdot[\bm{p}_{k+j|k}-\bm{p}_{k|k}]^\top + \zeta_i|}{\sqrt{\alpha_i^2+\beta_i^2+\gamma_i^2}} \right\} = 0.
\end{equation}
Likewise, constraint \eqref{eq:inputdelta_constraints} can be written as an equality constraint for $\phi$ and similarly for $\theta$ as follow:
\begin{subequations}\label{eq:delta_constraints}
\begin{align}
     \max\left\{0,\phi_{d,k+j|k} - \phi_{d,k+j+1|k}
    -\Delta \phi_{\max}\right\} =& 0,
    \\
   \max\left\{0, \phi_{d,k+j+1|k} - \phi_{d,k+j|k}
    -\Delta \phi_{\max}\right\} =& 0.
\end{align}
\end{subequations}

\section{Results} \label{sec:result}

\subsection{Simulation Setup}
The Gazebo robot simulator is used to evaluate the performance of the overall system architecture. The quad-copter in the simulations is equipped with the 3D lidar Velodyne VLP-16. During the simulations the ground truth odometry information provided from the Gazebo environment is used without considering any on-board sensor fusion. For evaluating the adaptive weights performance of the controller, random noises are added to the odometry measurements. We generate noise with the normal distribution $\mathcal{N}(\mu,\sigma^2)$~\cite{peebles2001probability}, where $\mu$ and $\sigma^2$ are the mean and standard deviation for each term of the states. From the practical point of view, position estimations suffer from higher uncertainties compared to velocity estimations~\cite{siegwart2011introduction}, since position drift is more difficult to recover compared to velocity drifts that can recover after few time steps. Thus, we consider higher standard deviation of noise for position estimations. The normal distributions, for the position and velocity estimation, are $\mathcal{N}(0,1.5)$ and $\mathcal{N}(0,0.5)$, respectively. 

Two environments are chosen to demonstrate the main attributes of the proposed navigation framework. The first environment is inspired from corridor areas, where the goal is to navigate from one end to the other, while avoiding collisions with the walls, the second case mainly focuses on the effect of odometry uncertainties in confined environments and demonstrates the performance when adaptively updating the uncertainty weights in the position and velocity in the controller. In all simulations, the term $\psi$ for the \gls{mav} is set to be zero because plane segmentation and \gls{nmpc} are in the body frame of \gls{mav} and independent of the yaw. The simulations can be found in the link: \url{https://youtu.be/76ob9HSrOAs}


The parameters of the \gls{mav} model are identical to the nonlinear model in~\cite{mansouri2020deploying}. 
Moreover, the tuning parameters of \gls{nmpc} are $\bm{Q}_{\Delta u}=[20,\,20,\,20]^\top$, $\bm{Q}_u=[10,\,10,\,10]^\top$, $T=[0,1]$, and $\phi_d,\,\theta_d =[-0.4, 0.4]~\unit{rad}$. 
The \gls{nmpc} prediction horizon $N$ is 40, the control sampling frequency $T_s$ is $\unit[20]{hz}$, $d_{s}$ sets to $\unit[1]{m}$, $n_{max}$ and $n_{cluster}$ are 10, and $\Delta \phi_{max}$ and $\Delta \theta_{max}$ are $\unit[0.05]{rad/sec}$. Moreover, the mean and maximum computation time of the proposed \gls{nmpc}, in all the studied scenarios, are $\unit[1.7]{msec}$ and $\unit[8.8]{msec}$ respectively. Finally, for the plane segmentation $\kappa_1$ is 0.1, $\kappa_2$ is 0.2, and $\lambda$ sets to 0.15, while the mean and maximum running time of the proposed segmentation method are $\unit[0.05]{sec}$ and $\unit[0.4]{msec}$. 




\subsection{Simulation Evaluation}

\subsubsection{Plane Segmentation}
\begin{figure*}[!htb]
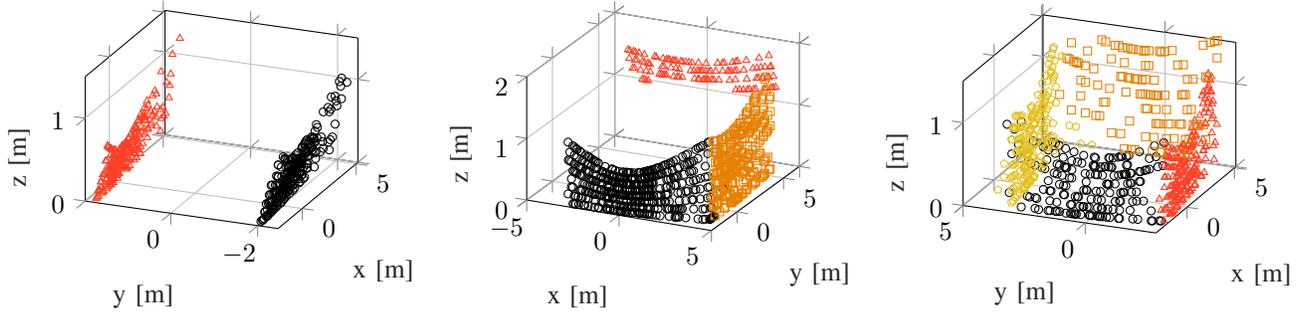

    \centering
    \setlength\fwidth{0.8\linewidth}
 \tikzsetnextfilename{gazeboenviroment} 
   \begin{tikzpicture}
   \setlength\fwidth{0.8\linewidth}
   \node at (0,0) [align=center] (left1) {\input{./Tikz/inclinewalls.tex}};
    \node[ right= 0.0 cm of left1,align=center] (middle1) {\input{./Tikz/ceilling.tex}};
    \node[ right= 0.0 cm of middle1,align=center] (right1) {\input{./Tikz/square.tex}};

 \end{tikzpicture}
    \caption{The extracted surfaces from various 3D point clouds in different scenarios based on the proposed segmentation method, each cluster is indicated with a different color. Best viewed in color.}
    \label{fig:planeresults}
\end{figure*}

\begin{figure}[!htb]
    \centering
    \setlength\fwidth{0.8\linewidth}
%
%
\definecolor{mycolor1}{rgb}{1.00000,0.25000,0.15000}%
\definecolor{mycolor2}{rgb}{0.90000,0.50000,0.00000}%
\definecolor{mycolor3}{rgb}{0.90000,0.75000,0.10000}%
\begin{tikzpicture}

\begin{axis}[%
width=0.951\fwidth,
height=0.4\fwidth,
at={(0\fwidth,1.1\fwidth)},
scale only axis,
xmin=95.00,
xmax=695.00,
xlabel style={font=\color{white!15!black}},
xlabel={Number of points},
ymode=log,
ymin=0.01,
ymax=8.00,
yminorticks=true,
ylabel style={font=\color{white!15!black}},
ylabel={Computation time [sec]},
axis background/.style={fill=white},
xmajorgrids,
ymajorgrids,
yminorgrids,
legend style={at={(0.0,1.041)}, anchor=south west, legend cell align=left, align=left, draw=white!15!black, legend columns=2}
]
\addplot [color=black, line width=1.0pt, mark size=2.0pt, mark=+, mark options={solid, black}]
  table[row sep=crcr]{%
100.00	0.01\\
200.00	0.03\\
300.00	0.03\\
400.00	0.05\\
600.00	0.07\\
700.00	0.07\\
};
\addlegendentry{Proposed method}

\addplot [color=mycolor1, line width=1.0pt, mark size=2.0pt, mark=o, mark options={solid, mycolor1}]
  table[row sep=crcr]{%
100.00	3.60\\
200.00	3.60\\
300.00	3.60\\
400.00	3.60\\
600.00	3.60\\
700.00	3.60\\
};
\addlegendentry{SSC-OMP}

\addplot [color=mycolor2, line width=1.0pt, mark size=2.0pt, mark=triangle, mark options={solid, mycolor2}]
  table[row sep=crcr]{%
100.00	6.70\\
200.00	6.70\\
300.00	6.70\\
400.00	6.70\\
600.00	6.70\\
700.00	6.70\\
};
\addlegendentry{SSC-ADMM}

\addplot [color=mycolor3, line width=1.0pt, mark size=2.0pt, mark=square, mark options={solid, mycolor3}]
  table[row sep=crcr]{%
100.00	2.60\\
200.00	2.60\\
300.00	2.60\\
400.00  2.60\\
600.00	2.60\\
700.00	2.60\\
};
\addlegendentry{ENSC}

\end{axis}

\end{tikzpicture}%
    \caption{Comparing running time of segmentation methods.}
    \label{fig:compareplanner}
\end{figure}
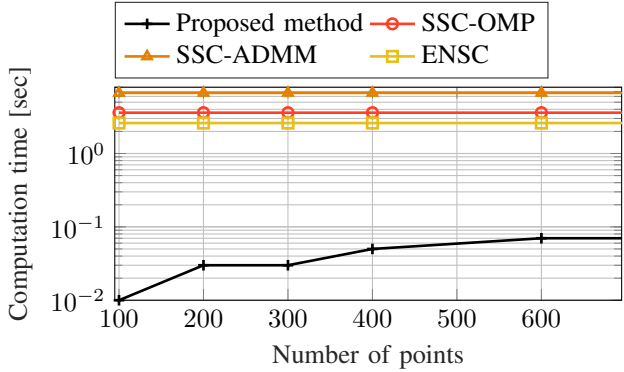

Figure~\ref{fig:planeresults} depicts the results of plane segmentation for different complex environments. The point clouds are in body coordinates, thus the \gls{mav} is located in the center. We also compared the running time of our method with \gls{ssc-omp}~\cite{you2016scalable}, \gls{ssc-admm}~\cite{elhamifar2013sparse}, and \gls{ensc}~\cite{you2016oracleoracle}. As depicted in Figure~\ref{fig:compareplanner}, our approach reduces the computational cost by at least an order of magnitude.

\subsubsection{Navigation in Corridor Environment}
\begin{figure*} [htbp!] \centering
 \includegraphics[width=0.8\linewidth, height=0.3\linewidth]{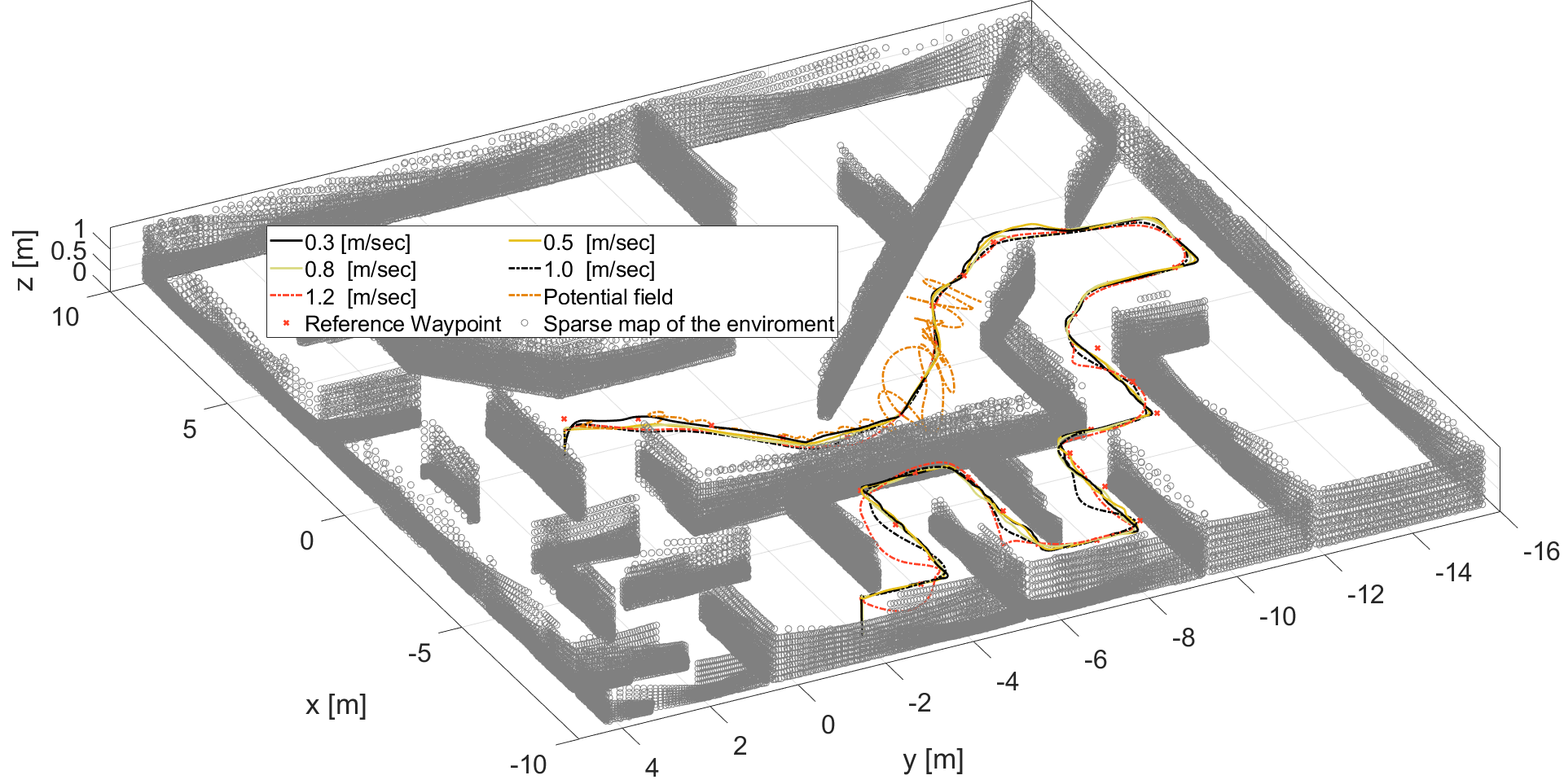}
 \caption{The trajectory of the \gls{mav} with different velocities during navigation in the corridor environment, while the way-points are shown by cross marker.}
 \label{fig:word1}
 \end{figure*}

Figure~\ref{fig:word1} depicts the trajectory of the \gls{mav} with the collision avoidance constraints of the \gls{nmpc} in the ``house\_maze'' world, a Gazebo environment available in VoxBlox repository\footnote{\url{https://github.com/ethz-asl/voxblox}}. The way-points are generated beforehand with different velocities, where some way-points have been intentionally selected close to the wall to demonstrate the collision avoidance capabilities of the framework in cases where the planner designs a trajectory prone to collisions with the environment.
This simulation does not include uncertainty linked with localization measurements. The proposed \gls{nmpc} controller avoids collisions for all velocities [0.3, 0.5, 0.8, 1.0, 1.2]~\unit[]{m/sec}. Additionally, the potential field method~\cite{droeschel2016multilayered} as a baseline local collision avoidance is evaluated with way-point generated with \unit[0.3]{m/sec}, while the same desired safety distance is set. The potential field trajectory is oscillating in the environment and it failed to reach the final goal. This is due to repulsive forces and large number of point clouds, which push the \gls{mav} from one side to another side.

The \gls{mae} between the desired trajectory and way-points are 0.16, 0,16, 0.22, 0.27, 0.36, and 1.4~\unit{m} for our proposed method with velocities of 0.3, 0.5, 0.8, 1.0, 1.2~\unit{m/sec} and potential field with velocity of 0.3\unit{m/sec} respectively. It should be highlighted that the \gls{mae} cannot be zero as some points are violating the constraints and the controller must avoid reaching them. It is observed that the \gls{mae} slightly increases with the increase in the generated way-point velocities. However, the increase in velocities did not result in any collision. The length of the trajectory for each velocity is 54.60, 50.60, 49.41, 49.68, 49.79~\unit{m} respectively. We see that the higher velocity results in slightly lower length of the path, however the \gls{mae} increases too. Nonetheless, this article is not focused on finding optimal path, and further investigation is required. Finally, the \gls{mae} of velocity profile of the \gls{mav} are 0.04, 0.03, 0.02, 0.12, 0.25~\unit{m/sec} for way-points of [0.3, 0.5, 0.8, 1.0, 1.2]~\unit[]{m/sec} velocities respectively. It is observed that the \gls{mae} is increased by increasing way-points velocities. However, it should be highlighted that in order to guarantee avoiding collision in confined environment, reaching higher speeds is not always a feasible solution. 

\begin{figure}[!htb]
    \centering
    \setlength\fwidth{0.8\linewidth}
\input{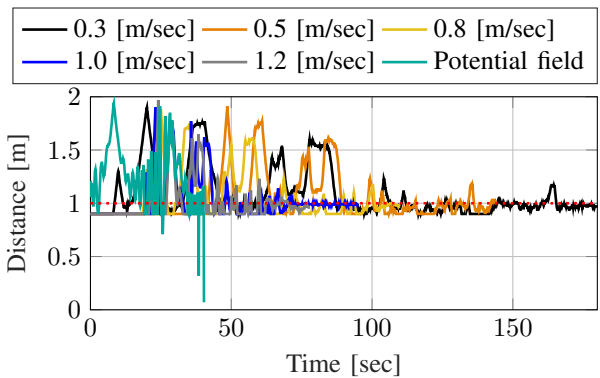}
    \caption{The minimum distance during navigation without localization uncertainties to the points in 3D point cloud, while the reference trajectory generated close to the obstacles.}
    \label{fig:word1distance}
\end{figure}

Figure~\ref{fig:word1distance} shows the minimum distance to the 3D point cloud during navigation without uncertainty in localization based on the proposed method with different velocities and the potential field method. It should be highlighted that the distance between confined areas of the environment is less than \unit[2]{m}, thus the constraints of $d_s=\unit[1]{m}$ is slightly violated with \unit[0.1]{m}. However, this does not have impact on overall performance and in all velocities we avoid collisions. 

\subsubsection{Navigation under Localization Uncertainties in Confined Environment}
\begin{figure} [htbp!] \centering
    \setlength\fwidth{0.8\linewidth}
\input{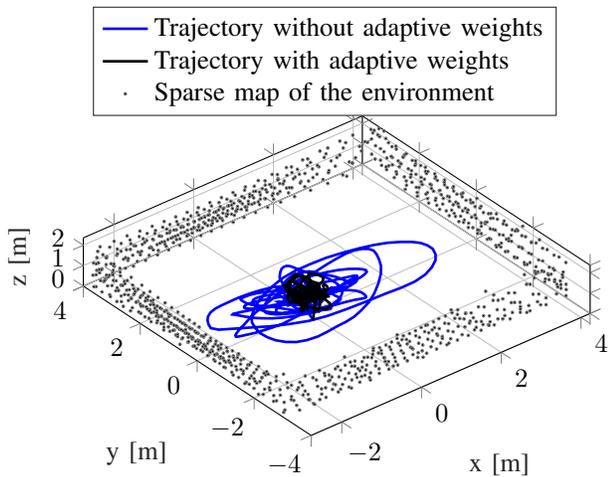}
\caption{The trajectory of the \gls{mav} during navigation in the confined environment, while the way-points sets to $[0,0,1,0,0,0]^\top$ and uncertainty in the localization is added.}
\label{fig:word2}
\end{figure}

\begin{figure}[!htb]
    \centering
    \setlength\fwidth{0.8\linewidth}
\input{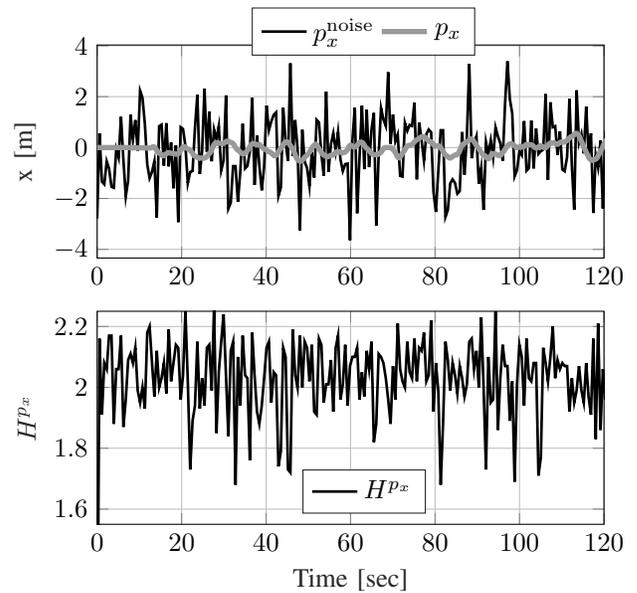}
    \caption{The real value and noisy measurement for position in $x$ axis and adaptive weight of $x$ tracking in the controller, the results are down sampled for better visualization.}
    \label{fig:odomuncertainty}
\end{figure}

\begin{figure}[!htb]
    \centering
    \setlength\fwidth{0.8\linewidth}
\input{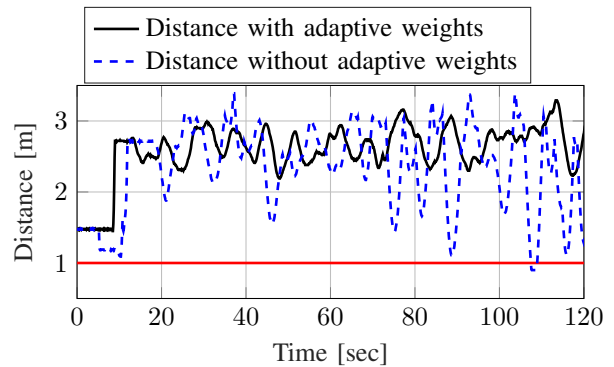}
    \caption{The minimum distance during  navigation with localization uncertainties to the points in 3D point cloud.}
    \label{fig:word2distance}
\end{figure}

 A confined environment with no entry/exit is chosen to evaluate the performance when uncertainties are induced in the position and velocity measurements in the $x$ and $y$ axis, while the reference way-point sets to $[0,0,1,0,0,0]^\top$. The noise in the measurements is induced after the take-off, when the \gls{mav} reaches the desired way-point. The focus of these simulations is to highlight the performance of the framework when the localization is noisy, which adaptively changes the weights in the position and the velocity based on the noise levels. 
Two scenarios are considered one with and one without adaptive weights for \gls{nmpc}, while in both cases the collision avoidance constraints are active. Figure~\ref{fig:word2} depicts the trajectory of the \gls{mav} in each scenario. This Figure shows that the \gls{mav} for the majority of the simulation run hovers close to the desired location when the adaptive weights are enabled, compared to the other case where it oscillates more. Figure~\ref{fig:odomuncertainty} shows the value of ground truth and measurements with noise for position in $x$ axis as an example, while the changes in the weights for tracking of $x$ is depicted too. It is observed that \gls{nmpc} with adaptive weights has less oscillation, and \gls{nmpc} without adaptive weights fails and collides to the walls. Moreover, Figure~\ref{fig:word2distance} shows the distance of each controller during the simulation, as it can be seen the adaptive weights of \gls{nmpc} improve the collision avoidance performance.


\section{Limitations and Future Improvements} \label{sec:limitations}

The proposed efficient planar segmentation method shows satisfactory performance in simulation results. However, false connections in the similarity graph constructed from sparse representations may lead to imprecise assignments, which is problematic for extracting planes from 3D point clouds. Additionally, the proposed method is mainly used for plane segmentation, and other shapes such as cylinders are not considered. Although the proposed approach is suitable for many robotic applications such as \gls{slam} and wall following, there is a need for more general object extraction techniques for local collision avoidance of \gls{mav} with \gls{nmpc}. Another future research direction is to analyze the effect of noisy point clouds on the accuracy of our plane segmentation method. 

\section{Conclusion} \label{sec:conclusions}
This article proposes a framework for autonomous navigation of \gls{mav}s in various environments with position uncertainties. The framework consists of two main modules, where the first one is plane segmentation from a local 3D point cloud based on a highly efficient sparse subspace clustering technique. The second module is \gls{nmpc} with collision avoidance based on plane equations and adaptive weights for tracking position, velocity or none based on localization uncertainties. The overall framework is evaluated in the Gazebo environment and the obtained results show the efficiency of the proposed methods to provide a collision-free navigation. It is shown that considering the localization uncertainties in controller results in robust maneuver, while without having adaptive weights, crushing of the \gls{mav} or larger drifts from desired trajectory are unavoidable.

 \bibliographystyle{IEEEtran}
%
\bibliography{mybib}
\end{document}